\documentclass[runningheads]{llncs}
\usepackage[T1]{fontenc}
\usepackage{graphicx}
\usepackage{subfigure}
\usepackage{booktabs}
\usepackage{ltablex}
\usepackage{xltabular}
\usepackage{pbox}
\usepackage{hyperref}

\usepackage{color}

\begin{document}
\title{Xplainer: From X-Ray Observations to Explainable Zero-Shot Diagnosis}

%
%
\author{Chantal Pellegrini\inst{1}\thanks{These authors contributed equally.},
Matthias Keicher\inst{1,\star}, Ege Özsoy\inst{1,\star}, Petra Jiraskova\inst{2}, Rickmer Braren\inst{2}, Nassir Navab\inst{1} }
\authorrunning{Pellegrini, Keicher and Özsoy et al.}
%
\institute{Computer Aided Medical Procedures, Technical University Munich, Germany \and Department of Diagnostic and Interventional Radiology, School of Medicine, Technical University of Munich, Germany}

\maketitle              
\begin{abstract}
Automated diagnosis prediction from medical images is a valuable resource to support clinical decision-making. However, such systems usually need to be trained on large amounts of annotated data, which often is scarce in the medical domain. Zero-shot methods address this challenge by allowing a flexible adaption to new settings with different clinical findings without relying on labeled data. Further, to integrate automated diagnosis in the clinical workflow, methods should be transparent and explainable, increasing medical professionals' trust and facilitating correctness verification. In this work, we introduce Xplainer, a novel framework for explainable zero-shot diagnosis in the clinical setting. Xplainer adapts the classification-by-description approach of contrastive vision-language models to the multi-label medical diagnosis task. Specifically, instead of directly predicting a diagnosis, we prompt the model to classify the existence of descriptive observations, which a radiologist would look for on an X-Ray scan, and use the descriptor probabilities to estimate the likelihood of a diagnosis. Our model is explainable by design, as the final diagnosis prediction is directly based on the prediction of the underlying descriptors. We evaluate Xplainer on two chest X-ray datasets, CheXpert and ChestX-ray14, and demonstrate its effectiveness in improving the performance and explainability of zero-shot diagnosis. Our results suggest that Xplainer provides a more detailed understanding of the decision-making process and can be a valuable tool for clinical diagnosis. Our code is available on github: \href{https://github.com/ChantalMP/Xplainer}{https://github.com/ChantalMP/Xplainer}
\end{abstract}

\section{Introduction}
Computer-aided diagnosis systems have become a prominent tool in medical diagnosis. Yet, their adoption is limited by the need for large amounts of annotated data for training, which hinders their scalability and adaptability to new clinical findings \cite{qin2018computer,fink2020potential}. Moreover, adapting to a new reporting template or clinical protocol necessitates new annotations, further reducing their feasibility in clinical settings. Recently, zero-shot \cite{huang2021gloria,boecking2022biovil,seibold2022breaking,tiu2022chexzero,wang2022medclip} and few-shot \cite{huang2021gloria,boecking2022biovil,keicher2022few} learning methods have been proposed as a potential solution, utilizing contrastive pretraining \cite{zhang2022contrastive,radford2021CLIP} on pairs of radiology reports and images, and achieving performance on par with radiologists \cite{tiu2022chexzero}. However, these methods lack the level of detail of radiology reports and inherent explainability, impeding their adoption in clinical settings \cite{kayser2022explaining}. Particularly, explaining the diagnosis with image descriptors is crucial to increase trust in the system and allow radiologists to verify the results \cite{mcinerney2023chill}.\looseness=-1
 
Inspired by the success of using large language models to predict image descriptors in natural images \cite{menon2022visual}, we introduce Xplainer, a novel framework that enhances the explainability of zero-shot diagnosis in the clinical setting. Xplainer leverages the classification-by-description approach \cite{menon2022visual} of vision-language models and adapts it to the multi-label medical diagnosis task. Specifically, we task the model to classify the existence of descriptive observations, which a radiologist would examine on an X-Ray scan, instead of directly predicting a diagnosis. This model design imbues our framework with intrinsic explainability, as the final diagnosis prediction is predicated on the underlying descriptor predictions.

We evaluate Xplainer on two chest X-ray datasets, CheXpert \cite{irvin2019chexpert} and ChestX-ray14 \cite{wang2017chestx}, and demonstrate its efficacy in enhancing the performance and explainability of zero-shot diagnosis in the clinical setting. Our results highlight that Xplainer provides a more comprehensive understanding of the diagnosis prediction process, thereby serving as a valuable tool for clinical decision-making. In summary, Xplainer presents a novel framework for zero-shot diagnosis that not only improves explainability and accuracy but also provides an invaluable tool for computer-aided diagnosis.
\section{Methodology}

\subsection{Model Overview}
We propose Xplainer, an explainable zero-shot classification-by-description approach for diagnosing pathologies from X-Ray scans. Given an image $i$ and a list of clinical observations $o_{p_{1-n}}$ per pathology $p$, the goal is to make a multi-label prediction indicating the diagnosis for the patient. 

Our zero-shot approach leverages the alignment of image and text embeddings provided by contrastive language-image pretraining (CLIP) \cite{radford2021CLIP}. We built upon BioVil \cite{boecking2022biovil}, a CLIP model pretrained on pairs of radiology reports and images. Employing the text and image encoders from BioVil, we calculate the cosine similarity between an X-ray image and each of $N$ pre-defined clinical observations $o_{p_{1-N}}$ describing a pathology. Then we calculate observation probabilities $P_{pos}(o_{p_{i}})$ for every observation. Analogously, we calculate probabilities for the absence of all observations $P_{neg}(o_{p_{i}})$ by defining negated prompts for all observations. Using the softmax over the positive and negative probability, we calculate the final probability of the presence of an observation $P(o_{p_{i}})$. Given these observation probabilities $P(o_{p_{i}}), i\in{1,...,N}$, we estimate a joined probability to determine the likelihood of the presence of a pathology $P(p)$:

\begin{equation}
    P(p) = \sum_{i=1}^{N} log(P(o_{p_{i}})) \div N
\end{equation}

\begin{figure}[tb]
\centering
\includegraphics[width=\textwidth]{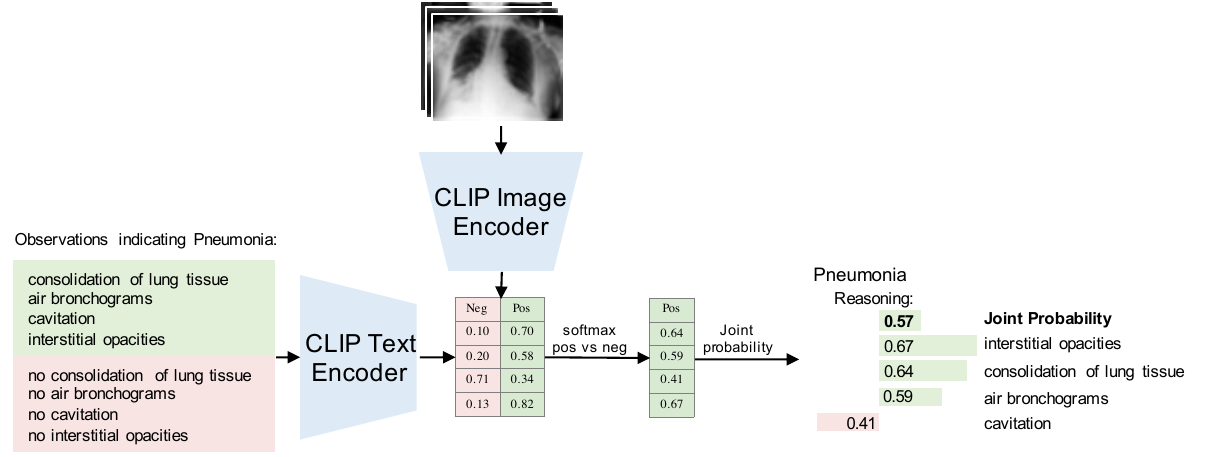}
\caption{Overview of Xplainer: In the first step, observation probabilities are calculated based on contrastive CLIP prompting. These are then used to make an explainable diagnosis prediction. The figure depicts an example for Pneumonia.}
\label{model_overview}
\end{figure}

\noindent We repeat this process for all pathologies we want to diagnose in the image. As the prediction of a pathology diagnosis is directly extracted from the observation probabilities, our method is explainable by design, producing a diagnosis prediction and the detected X-ray observations leading to that prediction. Moreover, the observation probabilities show which observations the model mainly considers for its diagnosis. Figure \ref{model_overview} shows an overview of our framework.

To integrate multiple images of one patient, we calculate positive and negative observation probabilities for each image and average them before calculating the pathology probability.

\subsection{Prompt Engineering}

Successful zero-shot inference relies on a good alignment between the contrastive pretraining and the downstream task \cite{radford2021CLIP}. As BioVil \cite{boecking2022biovil} was trained on pairs of radiological images and reports, we need to keep our observation prompts close to the style of medical reports. To initialize our prompts, we employ ChatGPT \cite{chatgpt} and query it to describe observations in X-ray images that would occur in a radiology report indicating specific pathologies. We further refined the prompts with the help of an experienced radiologist, who manually verified and adapted the descriptors. We provide a complete list of the descriptors in the supplementary.\looseness=-1

Radiology reports often include both presence and absence of particular observations. When comparing a prompt with an image embedding, it is hard for the model to differentiate between an observation's positive and negative occurrence, as their formulation can be very similar. Previous work \cite{seibold2022breaking,tiu2022chexzero} has shown that introducing negative prompts can circumvent this problem. Therefore, instead of thresholding the similarity between a positive prompt and an image, we prompt the model with both a positive and a negated version of each observation prompt and compare their probabilities. We adapt our prompts in two additional steps to align them with the text in radiology reports. First, we add a disease indication, as radiology reports usually contain observations paired with conclusions. Further, this reduces the ambiguity of our prompts, as in radiology, one sign (e.g., Lung Opacity) can indicate multiple pathologies (e.g., Pneumonia, Atelectasis, or Edema). Additionally, we frame all our observations in a sentence structure sounding more like an actual report by adding "There is/are" before every observation. Putting all of this together, we define the following prompt structure: "There is/are (no) <observation> indicating <pathology>."  Lastly, we define contrastive pathology-based prompts to compare to our observation-based prompting. In this setting, only two prompts, one positive and one negative prompt, are used per pathology. Overall, we compare the following styles of prompting to show the benefit of observation-based, contrastive prompting with disease indication and report style:

\noindent $\bullet$ \textbf{Pathology-based:} (No) <pathology>\\
$\bullet$ \textbf{Basic:} Only positive prompt per pathology: <observation>\\
$\bullet$ \textbf{Contrastive:} (No) <observation>\\
$\bullet$ \textbf{Pathology Indication:} (No) <observation> indicating <pathology>\\
$\bullet$ \textbf{Report Style:} There is/are (no) <observation> indicating <pathology>

\section{Experiments and Results}

We evaluate Xplainer in a zero-shot setting on the commonly used chest X-ray datasets, CheXpert \cite{irvin2019chexpert}, and ChestX-ray14 \cite{wang2017chestx}. The CheXpert dataset provides a manually labeled validation and test set with 200 and 500 patients, respectively, and 14 classes, including "No Finding", "Support Devices / Foreign Objects", and 12 pathology labels. ChestX-ray14 is evaluated on 14 pathology labels on a test set of 25.596 images. We perform a multi-label classification for both datasets and evaluate the performance via the Area Under the ROC-curve (AUC) between the positive pathology probabilities and the labels.

\begin{table}[b]
\centering
\caption{AUC for zero-shot pathology classification on CheXpert and ChestX-ray14 datasets. *in-domain, as the underlying CLIP model was trained the ChestX-ray14}
\begin{tabular}{|l|c|cc|c|}
\hline
             & CLIP pretraining data & \multicolumn{2}{c|}{CheXpert}     & ChestX-ray14 \\ \hline
             & & \multicolumn{1}{c|}{val}   & test & test         \\ \hline
             
CheXzero \cite{tiu2022chexzero} &  MIMIC & \multicolumn{1}{c|}{--} & 74.73    & --        \\ \hline
Seibold et al. \cite{seibold2022breaking} & MIMIC & \multicolumn{1}{c|}{78.86} & --    & 71.23        \\ \hline
Seibold et al. \cite{seibold2022breaking} & MIMIC, PadChest, ChestX-ray14  & \multicolumn{1}{c|}{83.24} & --    & 78.33*        \\ \hline
\textbf{Xplainer}  & MIMIC & \multicolumn{1}{c|}{\textbf{84.92}}  & \textbf{80.58} & \textbf{71.73}         \\ \hline
\end{tabular}
\label{tab:dataset_results}
\end{table}

\noindent Table \ref{tab:dataset_results} shows our results compared to previously proposed zero-shot pathology prediction approaches. On CheXpert, we compare with Seibold et al. \cite{seibold2022breaking} on the validation set, as they only reported validation performance. For the comparison with CheXzero \cite{tiu2022chexzero}, as well as the ChestX-ray14 dataset, we compare test set results. We outperform both previous works in an out-of-domain setting, where the zero-shot inference is performed on a different dataset than CLIP was trained on. The state-of-the-art results on both datasets show the effectiveness of our observation-based modeling. Further, in Table \ref{tab:pathology_results}, we provide a detailed breakdown of our results per pathology and dataset. 

\begin{table}[tb]
\centering
\caption{AUC per disease on both datasets}
\begin{tabular}{|l|c|c|c|}
\hline
                           & CheXpert Val & CheXpert Test & ChestX-ray14 \\ \hline
No Finding                 & 88.82         & 89.94          & --        \\ \hline
Enlarged Cardiomediastinum & 79.23         & 80.60          & --          \\ \hline
Cardiomegaly               & 78.62         & 83.32          & 79.71        \\ \hline
Lung Opacity               & 88.18         & 91.76          & --           \\ \hline
Lung Lesion                & 91.46         & 69.33          & --           \\ \hline
Edema                      & 84.84         & 84.55          & 81.46        \\ \hline
Consolidation              & 91.56         & 85.89          & 71.87        \\ \hline
Pneumonia                  & 85.68         & 83.73          & 70.83        \\ \hline
Atelectasis                & 84.64         & 85.46          & 66.86        \\ \hline
Pneumothorax               & 78.09         & 83.75          & 72.18        \\ \hline
Pleural Effusion           & 88.72         & 89.30          & 79.11        \\ \hline
Pleural Other              & 83.92         & 58.67          & --           \\ \hline
Fracture                   & --              & 60.47          & --           \\ \hline
Infiltration               & --            & --             & 68.81        \\ \hline
Mass                      & --            & --             & 70.28        \\ \hline
Nodule                     & --            & --             & 64.74        \\ \hline
Emphysema                   & --            & --             & 74.02        \\ \hline
Fibrosis                   & --            & --             & 62.25        \\ \hline
Pleural Thickening         & --            & --             & 67.44        \\ \hline
Hernia                     & --            & --             & 74.60        \\ \hline
Support Devices / Foreign Objects            & 80.25         & 81.15          & --           \\ \hline
\end{tabular}
\label{tab:pathology_results}
\end{table}

\subsubsection{Ablation Studies}
In our ablation studies, we investigate the impact of our prompt design and the effect of using multiple images. Table \ref{tab:prompt_comparison} shows the results on the CheXpert validation set using different prompting styles. We observe that pathology-based prompting, which reaches an AUC of 76.14\%, is significantly worse than observation-based prompting, which reaches an AUC of 84.92\%, again highlighting the benefit of observation-based prompting. Comparing the basic observation-based prompting, using only positive prompts per observation, to contrastive prompting, we see a substantial performance gap, showing the importance of using negative prompts to differentiate between positive and negative occurrences. We also show the effect of formulating our prompts unambiguously and in the style of an actual radiology report by adding pathology indication and report style. Adding pathology indication to the contrastive observation-based prompting significantly improves performance, achieving an AUC of 84.35\%. Finally, incorporating report style in the prompts leads to the highest AUC of 84.92\%, indicating that a contrastive observation-based prompt with pathology indication and report style is the most effective for zero-shot X-ray pathology classification.

\begin{table}[tb]
\centering
\caption{Comparison of different prompting styles on the validation set of CheXpert}
\begin{tabular}{|l|c|}
\hline
                   & AUC \\ \hline
Contrastive pathology-based Prompting       & 76.14  \\ \hline
\multicolumn{1}{|l}{\textbf{Observation-based Prompting:} } &      \\ \hline
Basic Prompt       & 58.65  \\ \hline
Contrastive Prompt & 77.00   \\ \hline
 $+$ pathology Indication & 84.35   \\ \hline
 $+$ Report Style       & 84.92   \\ \hline
\end{tabular}
\label{tab:prompt_comparison}
\end{table}

Additionally, we compare the initial ChatGPT output to our refined prompts (Table \ref{tab:prompt_refinement}). Refinement was performed by deleting irrelevant, redundant, or incorrect descriptors. We observe an improvement through the refinement, indicating that including domain knowledge further improves our method. Nevertheless, the original ChatGPT prompts already perform quite well, showing the impressive potential of combining large generic language models with large domain-specific contrastive models.

\begin{table}[tb]
\centering
\caption{Comparison of ChatGPT prompts vs. refinement with the help of a radiologist}
\begin{tabular}{|l|l|l|l|}
\hline
             & CheXpert Val & CheXpert Test & ChestX-ray14  \\ \hline
ChatGPT prompts & 83.61 & 79.94 & 71.40\\ \hline
Refined Prompts & \textbf{84.92} & \textbf{80.58} & \textbf{71.73}\\ \hline
\end{tabular}
\label{tab:prompt_refinement}
\end{table}

\noindent For the "No Finding" class, we compare to either define specific prompts such as "Clear lung fields" or "Normal heart size and shape" to classify "No Finding" or model it as the absence of all of the other 13 labels (Rule-based). Table \ref{tab:no_finding} shows that a rule-based modeling of this class leads to better results. A reason for this could be that there is no clearly defined list of observations that indicate a healthy X-ray scan, which a radiologist would mention in his report.

\begin{table}
\begin{minipage}[]{.45\textwidth}
\centering
\caption{Modeling of "No Finding" label with explicit prompts or rule-based definition as lack of other findings}
\begin{tabular}{lc}
\hline
\multicolumn{1}{|l}{}& \multicolumn{1}{|c|}{AUC - No Finding}  \\ \hline
\multicolumn{1}{|l}{Explicit Prompting}& \multicolumn{1}{|c|}{79.64} \\ \hline
\multicolumn{1}{|l}{Rule-based} & \multicolumn{1}{|c|}{\textbf{88.82}} \\ \hline
  & ~          \\
\end{tabular}
\label{tab:no_finding}
\end{minipage}\hfill
\begin{minipage}[]{.45\textwidth}
  \centering
\caption{Comparison of single-view inference to different methods for multi-image processing}
\begin{tabular}{|l|l|}
\hline
             & AUC  \\ \hline
Only single Frontal View & 84.19 \\ \hline
All - Max Aggregation          & 84.77 \\ \hline
All - Mean Aggregation      & \textbf{84.92} \\ \hline
\end{tabular}
\label{tab:multi_view}
\end{minipage}
\end{table}

\noindent Lastly, we investigate different image aggregation methods for pathology prediction. We compare only using a single frontal view X-ray to using all images available for a patient. For aggregation, we compute positive and negative observation probabilities for every image. In Max aggregation, we then use the highest observation probability. The intuition behind this approach is that an observation might be seen much better from one perspective than another, and then only the perspective where the model is most confident should be used. On the other hand, different views give different insights about which kind of observation a visual cue on the image indicates. To leverage this multi-view information, we test Mean aggregation, where all observation probabilities are averaged over multiple images. The results shown in Table \ref{tab:multi_view} indicate Mean aggregation to be superior, while both aggregation methods outperform using just a single image.

\subsubsection{Qualitative Results}
\begin{figure}[tb]
\centering
\includegraphics[width=\textwidth]{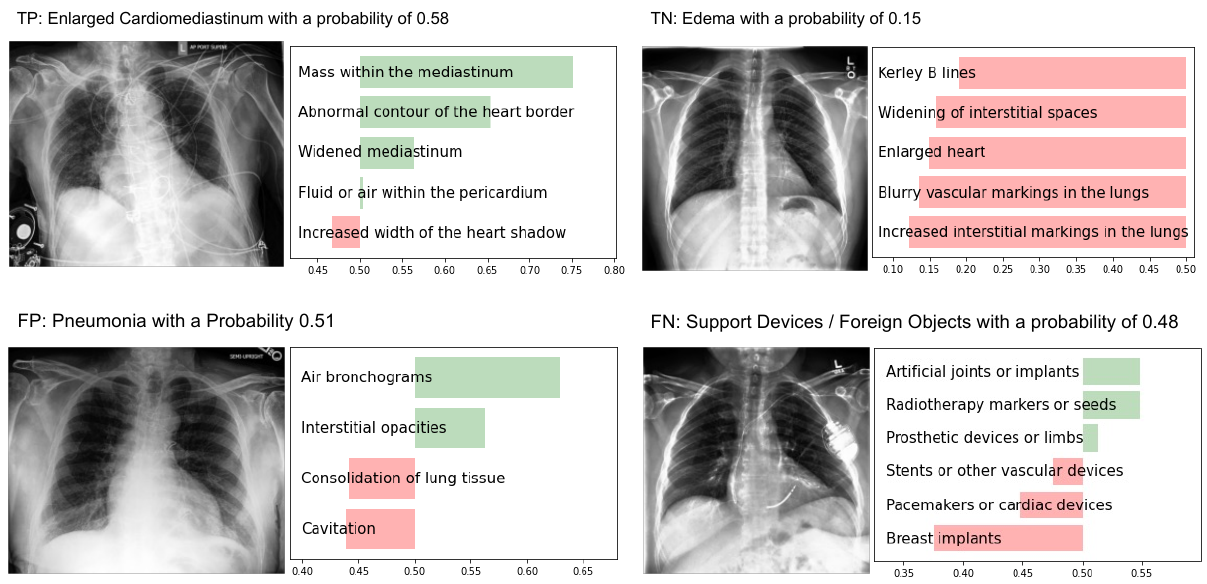}
\caption{Qualitative results of Xplainer}
\label{fig_qualitative}
\end{figure}

Figure \ref{fig_qualitative} shows qualitative examples of our model's predictions. For the true positive prediction, it can be seen that most of the descriptors are detected, and the model recognizes the descriptor "Mass in the mediastinum" as the main indication for the Enlarged Cardiomediastinum. For the True Negative case, the model, correctly, detected none of the descriptors. For the false positive example, one can clearly see that the model made a mistake because it detected an air bronchogram with relatively high certainty and no consolidation. Therefore, this false positive finding is easily falsified by the radiologist since an air bronchogram is a finding that co-occurs with consolidation (i.e., air-filled bronchi in consolidated areas). Thus, knowing which combination of descriptors leads to such a decision substantially improves explainability. In the false positive case, the model misses the pacemaker but detects some implant, showing the model understands there is some foreign object, but can not identify it, which is easily detected by the radiologist. Overall the classification-by-description may facilitate a plausibility check of a specific inference result and an understanding of the source of errors.

\subsubsection{Discussion}
One downside of modeling a joint probability is that it assumes that all descriptors appear simultaneously and gives all descriptors the same importance. While this estimation leads to good results, the assumption does not always hold, as a pathology does not always present with the same signs. Further, there might be inter-dependencies between the descriptors, e.g., there can be descriptors that strongly correlate with the presence of a disease when combined with one descriptor but much less when combined with another.
As a first try to model the importance of descriptors, we look into a supervised, out-of-domain approach to model these inter-dependencies. For this, we train a Naive Bayes~\cite{chan1982updating,Zhang2004} CheXpert classifier on MIMIC-CXR \cite{johnson2019mimic}, predicting a diagnosis given the descriptor probabilities, allowing the model to focus more on more relevant descriptors. While this approach relies on labels for MIMIC, these labels can be automatically generated by the CheXpert labeler \cite{irvin2019chexpert}, still not requiring human effort for labeling. We observe a slight performance increase on the test set from 80.58\% to 81.37\% AUC. This shows that the descriptor importance learned on MIMIC can partially be transferred to an out-of-domain dataset. We believe investigating methods to consider varying importance and complex relations between the descriptors is an essential and exciting direction to investigate in future work.\looseness=-1

The use of descriptors in Xplainer provides a flexible and adaptive approach to automated diagnosis prediction. By identifying and classifying the presence of descriptive observations, our model can capture the underlying characteristics of a disease without relying on labeled data. This means that our system can easily adapt to new settings with different clinical findings, including new conditions where the symptoms are known, but there is no training data available yet. Additionally, using descriptors allows for adapting the system to specific populations, where the essential descriptors can differ. This is because the model is not constrained by pre-defined labels but rather by the meaningful underlying features of a given diagnosis.
\section{Conclusion}
In this work, we present a novel and effective zero-shot approach for chest X-ray diagnosis prediction, which provides an explanation for the model's decision. We leverage BioVil, a pretrained, domain-specific CLIP model, and use contrastive observation-based prompting to make predictions without label supervision. Our approach significantly outperforms previous zero-shot methods on CheXpert and Chest-Xray14, showcasing the effectiveness of our approach. Furthermore, we show that designing informative prompts is crucial to improve model performance. Our ablation studies demonstrate that adding disease indication and report style formulation to observation-based prompts notably enhances performance, underscoring the importance of aligning prompts with the domain-specific language used in medical reports. Additionally, contrastive prompts significantly boost performance, suggesting that the model can benefit from explicitly contrasting positive and negative examples.

Our work highlights the potential of contrastive pretraining combined with observation-based prompting as a promising avenue for zero-shot medical image classification, where labeled data is scarce or expensive to obtain, and explainability is vital. We envision that our approach can be extended to other medical imaging domains and have practical applications in real-world scenarios. Our findings contribute to the growing body of research to improve the accuracy and interpretability of medical image diagnosis.

\section*{Acknowledgements}
The authors gratefully acknowledge the financial support by the Federal Ministry of Education and Research of Germany (BMBF) under project DIVA (FKZ 13GW0469C) and the Bavarian Research Foundation (BFS) under project PandeMIC (grant AZ-1429-20C).


%
%
%


\bibliographystyle{splncs04}
\bibliography{refs}
\section*{Supplementary}

{
\begin{xltabular}{\textwidth}{lX}
\caption{List of descriptors for all pathologies, separated by comma. An examplarory input for our method would be "There is \textit{increased size of the heart shadow} indicating \underline{enlarged cardiomediastinum}".}\\
\hline
No Finding & clear lung fields, normal heart size and shape, no abnormal fluid buildup, no visible tumors or masses, no signs of bone fractures or dislocations\\ \hline
\pbox{20cm}{Enlarged\\Cardiomediastinum} & increased width of the heart shadow, widened mediastinum, abnormal contour of the heart border, fluid or air within the pericardium, mass within the mediastinum  \\ \hline
Cardiomegaly & increased size of the heart shadow, enlargement of the heart silhouette, increased diameter of the heart border, increased cardiothoracic ratio \\ \hline
Lung Opacity & increased density in the lung field, whitish or grayish area in the lung field, obscured or blurred margins of the lung field, loss of normal lung markings within the opacity, air bronchograms within the opacity, fluid levels within the opacity, silhouette sign loss with adjacent structures  \\ \hline
Lung Lesion & consolidation of lung tissue, pleural effusion, cavities or abscesses in the lung, abnormal opacity or shadow in the lung, irregular or blurred margins of the lung  \\ \hline
Edema & blurry vascular markings in the lungs, enlarged heart, kerley b lines, increased interstitial markings in the lungs, widening of interstitial spaces  \\ \hline
Consolidation & loss of lung volume, increased density of lung tissue, obliteration of the diaphragmatic silhouette, presence of opacities \\ \hline
Pneumonia & consolidation of lung tissue, air bronchograms, cavitation, interstitial opacities\\ \hline
Atelectasis & increased opacity, volume loss of the affected lung region, blunting of the costophrenic angle, shifting of the mediastinum \\ \hline
Pneumothorax & tracheal deviation, deep sulcus sign, increased radiolucency, flattening of the hemidiaphragm, absence of lung markings, shifting of the mediastinum \\ \hline
Pleural Effusion & blunting of costophrenic angles, opacity in the lower lung fields, mediastinal shift, reduced lung volume, presence of meniscus sign or veil-like appearance \\ \hline
Pleural Other & pleural thickening, pleural calcification, pleural masses or nodules, pleural empyema, pleural fibrosis, pleural adhesions \\ \hline
Fracture & visible breaks in the continuity of the bone, misalignments of bone fragments, displacements of bone fragments, disruptions of the cortex or outer layer of the bone, visible callus or healing tissue, fracture lines that are jagged or irregular in shape, multiple fracture lines that intersect at different angles \\ \hline
\pbox{20cm}{Support Devices /\\ Foreign Objects} & artificial joints or implants, pacemakers or cardiac devices, stents or other vascular devices, prosthetic devices or limbs, breast implants, radiotherapy markers or seeds \\ \hline
Infiltration & irregular or fuzzy borders around white areas, blurring, hazy or cloudy areas, increased density or opacity of lung tissue, air bronchograms \\ \hline
Mass & calcifications or mineralizations, shadowing, distortion or compression of tissues, anomalous structure or irregularity in shape \\ \hline
Nodule & nodular shape that protrudes into a cavity or airway, distinct edges or borders, calcifications or speckled areas, small round oral shaped spots, white shadows \\ \hline
Emphysema & flattened hemidiaphragm, pulmonary bullae, hyperlucent lungs, horizontalisation of ribs, barrel chest \\ \hline
Fibrosis & reticular shadowing of the lung peripheries, volume loss, thickened and irregular interstitial markings, bronchial dilation, shaggy heart borders \\ \hline
Pleural Thickening & thickened pleural line, loss of sharpness of the mediastinal border, calcifications on the pleura, lobulated peripheral shadowing, loss of lung volume \\ \hline
Hernia & bulge or swelling in the abdominal wall, protrusion of intestine or other abdominal tissue, swelling or enlargement of the herniated sac or surrounding tissues, thickening of intestinal folds, retro-cardiac air-fluid level \\ \hline
\end{xltabular}
}
\end{document}